\documentclass[lettersize,journal]{IEEEtran}
\usepackage{amsmath,amsfonts}
\usepackage{algorithmic}
\usepackage{array}
\usepackage[caption=false,font=normalsize,labelfont=sf,textfont=sf]{subfig}
\usepackage{textcomp}
\usepackage{url}
\usepackage{verbatim}
\usepackage{graphicx}
\def\BibTeX{{\rm B\kern-.05em{\sc i\kern-.025em b}\kern-.08em
    T\kern-.1667em\lower.7ex\hbox{E}\kern-.125emX}}
\usepackage{balance}
\usepackage[normalem]{ulem}
\usepackage{amsmath}
\usepackage{amssymb}
\usepackage{mathtools}
\usepackage{amsthm}

\usepackage{dblfloatfix}
\usepackage{pifont}
\usepackage{microtype}
\usepackage{graphicx}
\usepackage{subcaption}
\usepackage{booktabs} 
\usepackage{multirow}  
\usepackage{tabularx}  
\usepackage{array}     
\usepackage[table]{xcolor} 

\usepackage{hyperref}
\usepackage[textsize=tiny]{todonotes}

\usepackage{cuted}
\usepackage{capt-of} 

\AtBeginDocument{%
  \providecommand\BibTeX{{%
    Bib\TeX}}}

\begin{document}
\title{Map2Vid: Embodied Navigation Video Generation \\with Human-Friendly Motion Control}

\author{Yulin~Dai, Yuqian~Fu, Tianwen~Qian,
Zhen~Xing, Bojia~Zi, and Xiaoling~Wang%
\thanks{Yulin Dai and Yuqian Fu contributed
equally to this work as the first authors.
(Corresponding authors: Tianwen Qian and
Xiaoling Wang.)}%
\thanks{Y. Dai, T. Qian, and X. Wang are with
East China Normal University,
Shanghai, China.}%
\thanks{Y. Fu is with King Abdullah University
of Science and Technology (KAUST),
Thuwal, Saudi Arabia.}%
\thanks{Z. Xing is with the Qwen Team,
Alibaba Group, Hangzhou, China.}%
\thanks{B. Zi is with The Chinese University
of Hong Kong, Hong Kong SAR, China.}%
}

\markboth{Journal of \LaTeX\ Class Files,~Vol.~18, No.~9, September~2020}%
{How to Use the IEEEtran \LaTeX \ Templates}

\IEEEaftertitletext{%
    \begin{minipage}{\textwidth}
        \centering
        \includegraphics[width=\linewidth]{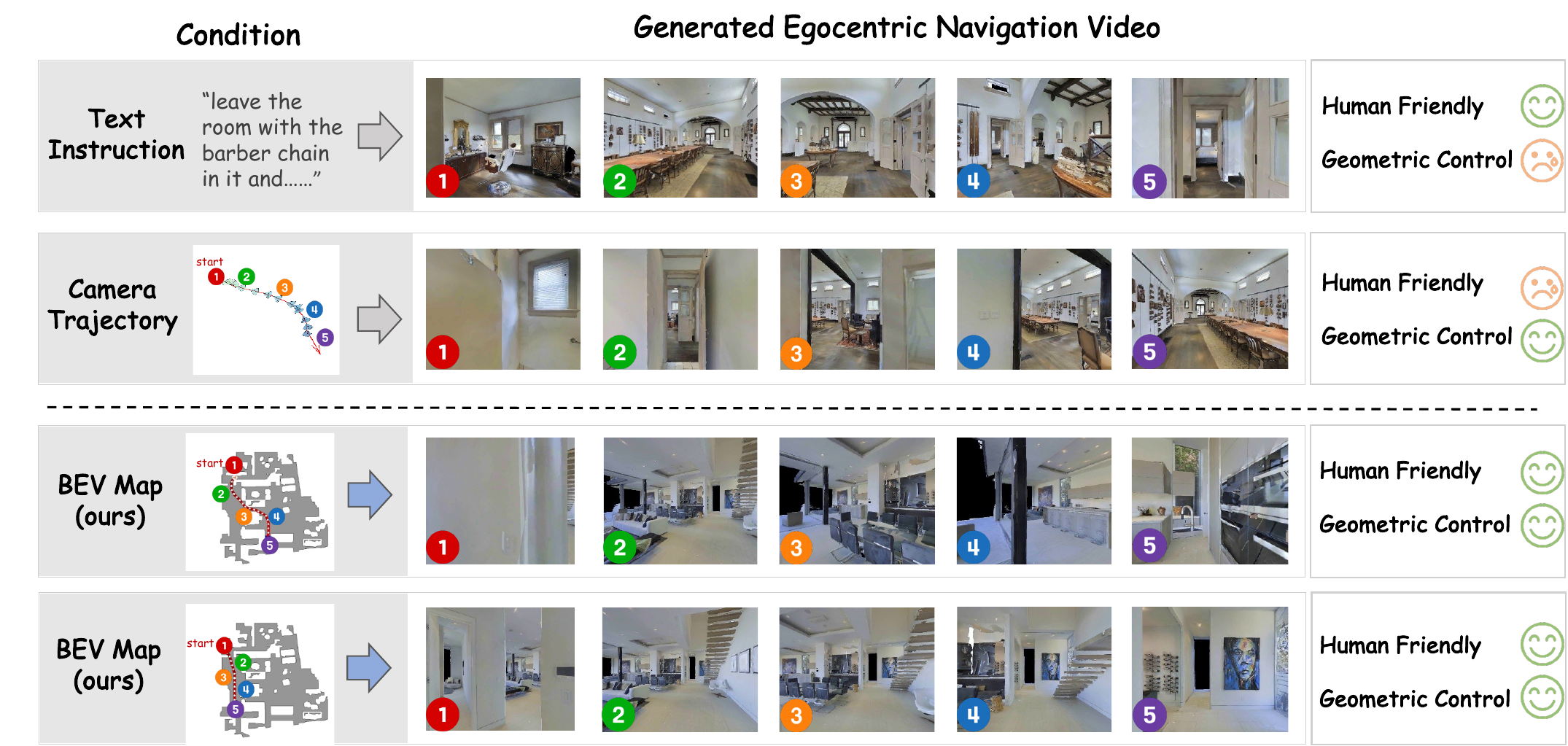}
        \captionof{figure}{
            \textbf{Comparison of control signals for embodied navigation video generation.}
            Existing methods rely on spatially ambiguous language (Top)
            or unintuitive camera trajectories (Middle).
            Map2Vid (Bottom) bridges this gap by conditioning on routed
            BEV map, combining human-friendly control with strong
            geometric control.
        }
        \label{fig:fig1}
    \end{minipage}%
}

\maketitle


\begin{abstract}
Controllable video generation has recently achieved impressive visual quality, yet existing control paradigms remain poorly suited to embodied navigation. Text prompts are spatially ambiguous, while camera-trajectory control relies on low-level motion specifications that are unintuitive for human users and difficult to manipulate. In this work, we introduce Map2Vid, a novel task that reformulates embodied navigation video generation as a BEV-map-to-video synthesis problem, together with a large-scale benchmark for systematic evaluation. By using BEV maps, which provide a visual representation of global scene layout and intended navigation routes, Map2Vid enables human-friendly navigation video generation. To support this task, we construct a dedicated dataset with over 100K navigation sequences across 61 indoor scenes, each paired with egocentric videos, BEV maps, camera trajectories and language instructions. We further develop a unified framework that integrates multiple control signals, together with a curriculum training strategy that progressively weakens non-map inputs, enabling flexible inference under both full conditioning and map-only input. Extensive experiments across multiple foundation generation models demonstrate that controllable navigation video generation via BEV map is a viable and promising direction.
\end{abstract}


\section{Introduction}


Embodied navigation~\cite{anderson2018evaluation, anderson2018vision, Yu2025MossVLN} requires an agent to perceive the world from an egocentric perspective and move through an environment under the geometric constraints imposed by scene layout. Beyond its importance to Embodied AI, this setting also exposes a distinctive challenge for controllable video generation: the synthesized video must remain temporally coherent~\cite{Zhao2026LSTD} under long-horizon ego-motion, while preserving scene geometry across turns, corridor traversals, and room transitions. Compared with generic text-to-video generation, navigation video synthesis therefore demands stronger spatial grounding, more reliable temporal consistency, and a control interface that can specify motion in an interpretable way.



Recent progress in large-scale video generation has substantially improved visual quality and motion realism. Most state-of-the-art approaches~\cite{wan2025wan, wu2025hunyuanvideo} condition the generation on text prompts and an initial frame, where textual descriptions primarily emphasize the appearance of the scene and cannot provide spatial grounding. This makes it difficult to maintain trajectory consistency and geometric plausibility over long horizons. Several recent works~\cite{he2024cameractrl,
wang2025ati} attempt to address this limitation by introducing explicit camera trajectory control.
Although effective in certain scenarios, they still suffer from two fundamental limitations: First, they support only local, small-scale camera motions (modest translations and rotations). In contrast, embodied navigation involves large-scale cumulative pose changes, including frequent large-angle turns and long-range transitions across rooms, resulting in a substantial mismatch in motion scale. Second, trajectory-based control depends on low-level specifications such as camera transformation matrices. Such representations offer limited abstraction over the spatial structure of the environment, and are fundamentally misaligned with how humans naturally express navigation intent, making them difficult for non-expert use.

These limitations motivate us to revisit the control signal itself. In natural navigation, humans rarely reason in terms of camera matrices; instead, they reason about the scene layout and the route through it .For this reason, we introduce \textbf{Map2Vid}, a new task that reformulates embodied navigation video generation as a BEV map. As shown in Fig.~\ref{fig:fig1}, Map2Vid takes as input a BEV map depicting the scene layout together with an intended navigation route, and aims to generate an egocentric video consistent with the specified path. Compared to text prompt or camera trajectory, BEV map provides 
structurally grounded and globally consistent control signals,
enabling a more human-friendly control interface for navigation video generation. To instantiate this task, we construct a large-scale dataset \textbf{Map2Vid-100K}. Specifically, we leverage high-fidelity indoor environments of Matterport3D~\cite{ramakrishnan2021hm3d}, and sample valid navigation trajectories from the R2R benchmark~\cite{anderson2018vision}. Each navigation episode is executed in the Habitat-Sim~\cite{puig2023habitat3} simulator, during which we record the corresponding egocentric video together with aligned natural language instruction, camera trajectories and BEV map. The dataset covers 61 indoor scenes and contains 117.8K navigation sequences, with 94.2K used for training and 23.6K reserved for evaluation. While our benchmark is constructed in simulation, it provides precise geometric alignment across modalities, which is difficult to obtain in real-world settings.

Building on this dataset, we adapt several representative video generation models to support map-conditioned navigation video synthesis. During training, models are provided with multiple control signals, including the BEV map, camera trajectories, text instruction, and initial frame. To prevent models from over-relying on expressive but less human-friendly controls, we propose a curriculum training strategy that gradually weakens non-map conditions, encouraging the model to internalize the spatial dynamics semantics encoded in the BEV map. This design enables flexible inference at test time, supporting either full control signals or the BEV map alone. Extensive experiments demonstrate that controllable navigation video generation via BEV map is viable and promising, while also highlighting the intrinsic difficulty of the task.

In summary, our contributions are threefold: (1) We introduce Map2Vid, a novel task that reformulates embodied navigation video generation as a BEV map, providing a human-friendly control interface. (2) We construct Map2Vid-100K, a large-scale benchmark comprising over 117K synchronized multimodal navigation sequences (videos, BEV maps, trajectories, instructions) across 61 scenes, establishing a standardized dataset and dual-aspect evaluation protocol for this new task. (3) We provide a systematic empirical study across four state-of-the-art video foundation models and propose a unified baseline framework with a progressive curriculum training strategy, validating the feasibility of the task and laying a strong empirical foundation for future research.




\section{Related Work}
\label{sec:related_work}

\begin{figure*}
    \centering
    \includegraphics[width=0.9\textwidth]{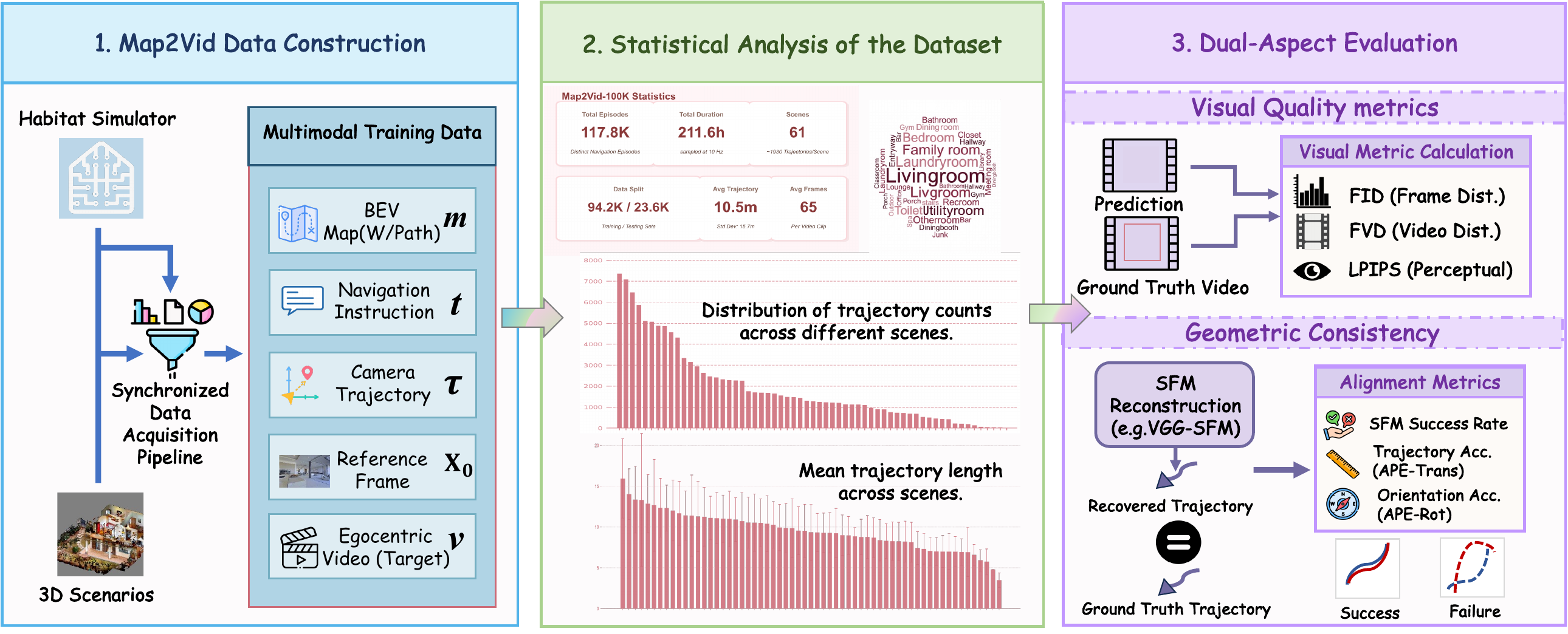}
    \caption{\textbf{Overview of the benchmark.} 
    It consists of three components: 
    (1) \textbf{Data Construction}: synchronized multimodal data collection in the Habitat simulator. 
    (2) \textbf{Dataset Statistics}: covering diverse scene semantics and a wide range of navigation distances, highlighting both the diversity and scale of the benchmark.
    (3) \textbf{Evaluation Metrics}: comprehensive assessment along two complementary dimensions, visual fidelity and geometric consistency.}
    \label{fig:fig2}
\end{figure*}

\textbf{Controllable Video Generation.}
Recent advances in video generation have been significantly driven by scalable Diffusion Transformers (DiT)~\cite{peebles2023scalable}. Prominent examples of this architecture include flagship models such as HunyuanVideo~\cite{wu2025hunyuanvideo, kong2024hunyuanvideo} and Wan~\cite{wan2025wan}.
Despite these successes, achieving fine-grained controllability in video generation remains challenging. Early approaches adapted image-based conditioning techniques, such as ControlNet~\cite{zhang2023adding} and T2I-Adapter~\cite{mou2024t2i}, to the temporal dimension for structural guidance~\cite{guo2023animatediff, wang2023videocomposer, hu2024animate}. While these methods effectively enforce spatial structures (e.g., Canny edges, depth maps, or pose skeletons), they often struggle to maintain long-horizon temporal coherence and 3D geometric consistency.
To better control motion dynamics, subsequent works (e.g., MotionCtrl~\cite{wang2024motionctrl}, DragNUWA~\cite{yin2023dragnuwa} and CameraCtrl~\cite{he2024cameractrl}) introduce explicit camera trajectory-based motion conditioning, with CameraCtrl~\cite{he2024cameractrl} standing out with its Plucker embeddings. Though these methods enable the geometry control of the video generation, the input form of accurate camera trajectories is too non-trivial for normal users. By contrast, in this paper, we for the first time introduce and explore the BEV map as a geometric condition.


\noindent
\textbf{Embodied Video Understanding \& Synthesis.}
\label{subsec:embodied_synthesis}
Advanced by large-scale egocentric datasets such as Ego4D~\cite{grauman2022ego4d} and Ego-Exo4D~\cite{grauman2024ego}, the field has expanded from passive video understanding to agent-centric embodied world models (e.g., GAIA-1~\cite{hu2023gaia}, UniSim~\cite{yang2023unisim}). In the realm of embodied video synthesis, existing works primarily address geometric consistency through trajectory conditioning in structured driving domains~\cite{gao2023magicdrive,wang2024drivedreamer} or via cross-view synthesis from third-person observations~\cite{liu2024exocentric}. Distinct from these approaches, the potential of top-down BEV maps as a spatially grounded control signal remains largely underexplored.
Embodied AI has increasingly shifted from passive perception toward agent-centric understanding, supported by large-scale egocentric datasets such as Ego4D~\cite{grauman2022ego4d} and Ego-Exo4D~\cite{grauman2024ego}. Beyond understanding, recent work has explored video generation as a foundation for embodied world models that simulate agent–environment interactions, exemplified by GAIA-1~\cite{hu2023gaia} and UniSim~\cite{yang2023unisim}. A central challenge in this line of research is generating egocentric video that remains consistent with agent motion and scene geometry over long horizons. Several studies address this challenge through trajectory-conditioned generation, particularly in structured domains such as autonomous driving, where motion patterns and scene layouts are relatively constrained (e.g., DriveDreamer~\cite{wang2024drivedreamer}, MagicDrive~\cite{gao2023magicdrive}). Other works focus on alleviating the scarcity of egocentric data by synthesizing egocentric views from third-person observations (Exo2Ego), using multi-view and diffusion-based frameworks~\cite{liu2024exocentric}.


\section{Map2Vid: Setup, Benchmark, and Metrics}
\label{sec:task}
\subsection{Task Formulation}
\label{sec:task_setup}
The primary objective of the Map2Vid task is to enable the synthesis of embodied egocentric navigation videos conditioned on a BEV map, supplementing conventional modalities such as images and text, as well as the recently explored camera trajectory modality.
Formally, let $\mathcal{V} \in \mathbb{R}^{T \times H \times W \times 3}$ denote a target video sequence of length $T$, representing an agent's first-person view navigating through a scene. We define the conditional generation problem as modeling the probability distribution $p(\mathcal{V} | \mathcal{C})$, where $\mathcal{C}$ represents the set of control conditions. 
To fully leverage all available and complementary signals, we define $\mathcal{C}$ to include the following four modalities:

\begin{enumerate}
    \item \textbf{BEV Map ($\mathbf{m}$):} A bird's-eye-view image of the environment, overlaid with a graphical depiction of the intended navigation route.
    \item \textbf{Navigation Instruction ($\mathbf{q}$):} A natural language description of the path (e.g.,``Walk past the kitchen...").
    \item \textbf{Camera Trajectory ($\boldsymbol{\tau}$):} A sequence of camera parameters $\boldsymbol{\tau} = \{\mathbf{p}_t\}_{t=1}^T$, where each $\mathbf{p}_t \in \mathbb{R}^{16}$ fully parameterizes the projection model of the $t$-th frame. Specifically, $\mathbf{p}_t$ is formed by concatenating the intrinsic parameter vector $\mathbf{k} = [f_x, f_y, c_x, c_y]^\top$ and the row-major flattened extrinsic matrix $M_{\text{ext}} = [R | \mathbf{t}] \in \mathbb{R}^{3 \times 4}$. 
    \item \textbf{Reference Frame ($\mathbf{x}_0$)}: 
    The initial observation.
\end{enumerate}

\noindent
\textbf{Training Phase:} 
To maximize the model's capability, we allow access to full set of conditions $\mathcal{C}$ during the training stage. Thus, the objective is defined as:
\begin{equation}
    \mathcal{L}_{train} = \mathbb{E}_{\mathcal{V}, \mathbf{m}, \mathbf{q}, \boldsymbol{\tau}, \mathbf{x}_0} [\log p_\theta(\mathcal{V} | \mathbf{m}, \mathbf{q}, \boldsymbol{\tau}, \mathbf{x}_0)].
\end{equation}

\noindent
\textbf{Inference Phase:} 
Focusing on the BEV map condition, we consider two scenarios during inference: one using the full condition set $\{\mathbf{m}, \mathbf{q}, \boldsymbol{\tau}, \mathbf{x}_0\}$ (Eq.~\ref{eq:test1}), and one conditioned exclusively on the BEV map $\mathbf{m}$ (Eq.~\ref{eq:test2}):
\begin{equation}\label{eq:test1}
    \hat{\mathcal{V}} \sim p_\theta(\mathcal{V} | \mathbf{m}, \mathbf{q}, \boldsymbol{\tau}, \mathbf{x}_0).
\end{equation}
\begin{equation}\label{eq:test2}
    \hat{\mathcal{V}} \sim p_\theta(\mathcal{V} | \mathbf{m}).
\end{equation}

\subsection{Map2Vid Benchmark}
\label{sec:task_bench}

To facilitate the Map2Vid task, we introduce Map2Vid-100K, a large-scale multimodal dataset consisting of aligned egocentric videos, BEV maps, language instructions, camera trajectories, and reference frames. It is constructed by extending the Room-to-Room (R2R) benchmark~\cite{anderson2018vision} within the Habitat simulator~\cite{puig2023habitat3} through two steps:

\noindent
\textbf{\ding{172} Scene Curation and Trajectory Generation:}
The R2R benchmark provides a large collection of language-guided navigation trajectories grounded in Matterport3D~\cite{ramakrishnan2021hm3d} environments, making it well suited for embodied navigation data construction. Specifically, we employ the Habitat simulator to re-simulate the training split of the R2R dataset, preserving the original navigation instructions and action sequences.
To support our scene-by-scene fine-tuning setting, we treat different floors within the same scan as independent scenes and discard trajectories involving cross-floor transitions, ensuring single-level geometric consistency. Within each curated scene, an agent executes the ground-truth actions to generate navigation episodes.

\noindent
\textbf{\ding{173} Synchronized Data Acquisition:} 
During execution, we synchronously record the multimodal data tuple $\{\mathcal{V}, \mathbf{m}, \mathbf{q}, \boldsymbol{\tau}, \mathbf{x}_0\}$. Specifically, camera trajectories $\boldsymbol{\tau}$ are obtained by logging agent kinematics. The egocentric videos $\mathcal{V}$ and initial frames $\mathbf{x}_0$ are rendered based on $\boldsymbol{\tau}$, while the text instructions $\mathbf{q}$ are inherited from R2R annotations. Finally, the BEV maps $\mathbf{m}$ are generated by projecting the 3D scene geometry into 2D occupancy maps and overlaying the executed navigation routes as a graphical trajectory.

\noindent
\textbf{Benchmark Statistics.}
As illustrated in Fig.~\ref{fig:fig2}, Map2Vid-100K comprises 117.8K navigation episodes spanning 61 indoor scenes, with 94.2K for training and 23.6K for testing. The generated egocentric videos have an average of 65 frames, sampled at 10 Hz, amounting to 211.6 hours of video in total. Trajectories have an average length of 10.5 meters (Std. 15.7 m), covering short segments to long-horizon paths with substantial ego-motion.

\subsection{Evaluation Metrics}
\label{sec:task_metrics}
To evaluate the generated videos comprehensively, we propose a series of evaluation metrics, encompassing both visual quality and geometric consistency assessment.

\noindent
\textbf{Visual Quality Metrics.}
We report standard video generation metrics to assess the realism and perceptual quality. Specifically, \textit{Fr\'echet Video Distance (FVD)}~\cite{unterthiner2018towards} measures distributional discrepancy at the temporal level, and \textit{Fr\'echet Inception Distance (FID)}~\cite{heusel2017gans} evaluates the frame level visual quality. In addition, \textit{LPIPS}~\cite{zhang2018unreasonable} is used to quantify the perceptual similarity.

\noindent
\textbf{Geometric Consistencies.}
Standard metrics are insufficient for embodied navigation videos, as they fail to capture whether an agent follows the intended path or respects the scene geometry. To address this limitation, we propose a reference-free evaluation pipeline that explicitly measures the geometric consistency between the conditioned trajectory $\boldsymbol{\tau}$ and the camera motion exhibited in the generated video $\hat{\mathcal{V}}$.
Specifically, we apply \textbf{VGG-SfM}~\cite{wang2024vggsfm} to recover an estimated camera trajectory $\hat{\boldsymbol{\tau}} = \{\hat{\mathbf{p}}_t\}_{t=1}^T$ from $\hat{\mathcal{V}}$, and then compare it with the ground truth trajectory $\boldsymbol{\tau} = \{\mathbf{p}_t\}_{t=1}^T$. 
Based on the reconstructed trajectories, we define three specific metrics:

\begin{itemize}
    \item \textbf{SfM Success Rate (SfM-SR):} This metric quantifies the structural integrity of the generated videos. We define it as the proportion of samples for which SfM successfully produces a valid reconstruction:
    \begin{equation}
        \text{SfM-SR} = \frac{1}{N} \sum_{i=1}^{N} \mathbb{I}(\text{SfM}(\hat{\mathcal{V}}_i) \neq \emptyset),
    \end{equation}
    where $\mathbb{I}(\cdot)$ is the indicator function and $\text{SfM}(\cdot) \neq \emptyset$ denotes a successful reconstruction. Failures typically indicate severe temporal flickering or geometric distortions that prevent reliable feature tracking.
    
    \item \textbf{Trajectory Accuracy ($\mathbf{APE}_{\text{trans}}$):} Following~\cite{grupp2017evo}, we compute the absolute pose error (APE) for translation. After aligning $\hat{\boldsymbol{\tau}}$ to $\boldsymbol{\tau}$ via Umeyama alignment, the translational error is measured as:
    \begin{equation}
        \text{APE}_{\text{trans}} = \sqrt{\frac{1}{T} \sum_{t=1}^{T} \| \mathbf{t}_{t} - \hat{\mathbf{t}}'_{t} \|_2^2},
    \end{equation}
    where $\mathbf{t}_{t}$ and $\hat{\mathbf{t}}'_{t}$ denote the ground truth and the estimated translation, respectively.
    
    \item \textbf{Orientation Accuracy ($\mathbf{APE}_{\text{rot}}$):} We evaluate rotational accuracy as the mean angular error between aligned rotation matrices:
    \begin{equation}
        \text{APE}_{\text{rot}} = \frac{1}{T} \sum_{t=1}^{T} \left| \angle(\mathbf{R}_{t} (\hat{\mathbf{R}}'_{t})^{\top}) \right|,
    \end{equation}
    where $\mathbf{R}_{t}$ and $\hat{\mathbf{R}}'_{t}$ denote the ground-truth and aligned estimated rotations, respectively. $\angle(\cdot)$ computes the rotation angle of the residual rotation matrix.
\end{itemize}

\begin{figure*}[ht]
    \centering
    \includegraphics[width=0.9\textwidth]{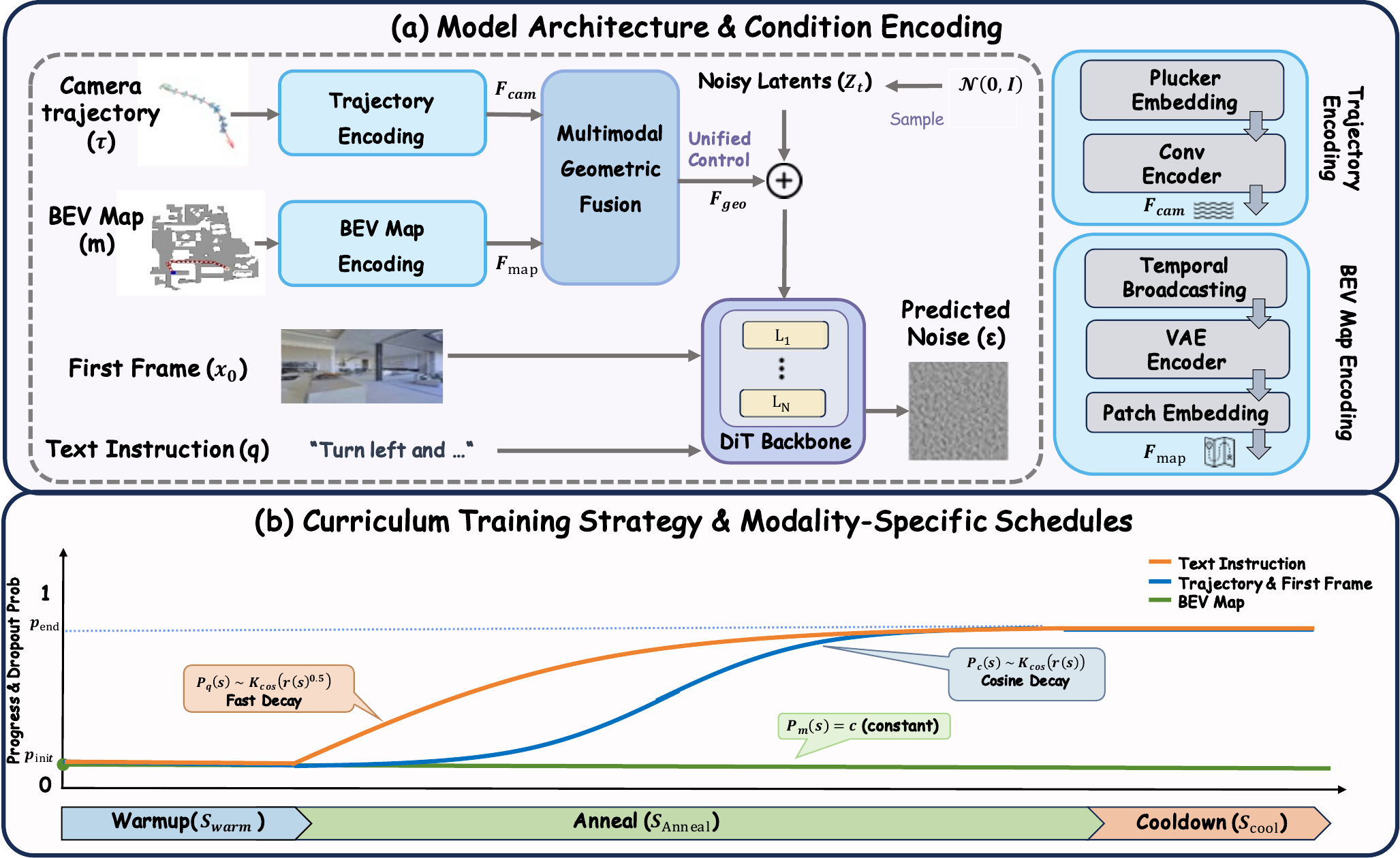}
    \caption{\textbf{Overview of the framework and training strategy.} 
    \textbf{(Top) Architecture:} We adopt a hybrid condition injection on a DiT-based backbone. 
    Instruction $\mathbf{q}$ and the first frame $\mathbf{x}_0$ are injected through the model’s native pathways, while camera trajectory $\boldsymbol{\tau}$ and BEV map $\mathbf{m}$ are encoded via Pl\"{u}cker embeddings and temporal broadcasting with a VAE Encoder, respectively, and then fused via a fusion layer.
    \textbf{(Bottom) Training:} A progressive masking schedule gradually weakens non-map conditions, with modality-specific designs.}
    \label{fig:overview}
\end{figure*}

\section{Methodology} \label{sec:method}

\subsection{Preliminaries \& Overview}
\noindent \textbf{Preliminaries.}
We build upon DiT-based latent video diffusion models~\cite{rombach2022high}, which have become a standard paradigm for high-fidelity video generation. Let $\mathcal{V}$ denote a navigation video and $z_0 = \mathcal{E}(\mathcal{V})$ be its latent representation encoded by a pre-trained encoder $\mathcal{E}$ (e.g., a VAE~\cite{kingma2013auto}). At diffusion step $t$, Gaussian noise $\epsilon$ is added to obtain the $z_t$. Existing models typically condition the denoising network $\epsilon_\theta$ on a text instruction $\mathbf{q}$ and a reference frame $\mathbf{x}_0$, and the objective is:
\begin{equation}
\mathcal{L}_{\text{base}} = \mathbb{E}_{z_0, \epsilon, \mathbf{q},  \mathbf{x}_0} \left[ \| \epsilon - \epsilon_\theta(z_t, \mathbf{q}, \mathbf{x}_0) \|_2^2 \right],
\label{eq:base_loss}
\end{equation}
where $\epsilon$ denotes the ground-truth noise.
To support Map2Vid, we extend the conditioning framework to incorporate motion signals: the BEV map $\mathbf{m}$ and camera trajectory $\boldsymbol{\tau}$. Consequently, our objective is formulated as:
\begin{equation}
\mathcal{L}_{\text{ours}} = \mathbb{E}_{z_0, \epsilon, \mathcal{C}} \left[ \| \epsilon - \epsilon_\theta(z_t, \underbrace{\mathbf{q}, \mathbf{x}_0}_{\text{Native}}, \underbrace{\boldsymbol{\tau}, \mathbf{m}}_{\text{Geometric}}) \|_2^2 \right],\label{eq:our_loss}
\end{equation}
where $\mathcal{C}$ denotes the full set of conditions, i.e.,  $\mathcal{C} = \{\mathbf{m}, \mathbf{q}, \boldsymbol{\tau}, \mathbf{x}_0\}
$.

\noindent
\textbf{Framework Overview.} As illustrated in Fig.~\ref{fig:overview}, we extend the original backbone with geometric conditions injection. During training, the model is conditioned on the full set of control signals. Additionally, we employ a curriculum training scheme that progressively weakens non-map conditions, encouraging the model to internalize map-centric information and enabling flexible inference with either full conditioning or map-only input. Details of the geometric condition injection and the curriculum training are presented below.

\subsection{Geometric Conditions Injection}

\textbf{Pl\"{u}cker Embedding for Camera Trajectory.}
Directly concatenating raw camera matrices (extrinsics and intrinsics) to visual features is suboptimal due to the high dimensionality and non-spatial nature of the parameters. Inspired by CameraCtrl~\cite{he2024cameractrl}, we leverage Pl\"{u}cker coordinates to explicitly model the ray geometry of the camera sensor.
Concretely, for each step $t$ in $\boldsymbol{\tau} = \{\mathbf{p}_t\}_{t=1}^T$, the camera parameters $\mathbf{p}_t \in \mathbb{R}^{16}$ are mapped to a 6-channel Pl\"{u}cker embedding map $\mathbf{E}_{\text{Pl\"{u}cker}} \in \mathbb{R}^{T \times 6 \times H \times W}$. Each pixel $(u, v)$ in the feature map is represented by a ray $(\mathbf{o}, \mathbf{d})$, where $\mathbf{o}$ is the ray origin and $\mathbf{d}$ is the ray direction. This representation preserves the 3D geometric inductive bias. 
This embedding is then processed by a lightweight convolutional encoder $\phi_{cam}$ to yield the camera control feature:
\begin{equation}
\mathbf{F}_{cam} = \phi_{cam}(\mathbf{E}_{\text{Pl\"{u}cker}}).
\end{equation}

\noindent
\textbf{Spatio-Temporal Embedding for BEV Map.}
The BEV $\mathbf{m}$ provides global spatial context but lacks a temporal dimension. To align it with the video generation, we treat the map as a static video sequence and process it via three steps:

\begin{enumerate}
    \item \textbf{Temporal Broadcasting:} The static map image $\mathbf{m}$ is replicated $T$ times to match the video length, forming a pseudo-map video sequence $\mathbf{m}_{seq} \in \mathbb{R}^{T \times 3 \times H \times W}$.
    \item \textbf{Latent Compression:} To match the distribution of the target video latents, we encode $\mathbf{m}_{seq}$ using the same visual encoder $\mathcal{E}$ used for the video $\mathcal{V}$, resulting in latent map features $\mathcal{E}(\mathbf{m}_{seq})$.
    \item \textbf{Patch Embedding:} Since the map is now structured in a video-like format, we tokenize the latents $\mathcal{E}(\mathbf{m}_{seq})$ using standard patch embedding layers~\cite{dosovitskiy2020image} to obtain the feature map $\mathbf{F}_{map}$.
\end{enumerate}

\noindent
\textbf{Multimodal Geometric Conditions Fusion.}
Recognizing that both the BEV map and the camera trajectory describe the spatial geometry of the scene, we propose to fuse these complementary signals. Specifically, we first concatenate $\mathbf{F}_{map}$ and $\mathbf{F}_{cam}$ along the channel dimension and subsequently capture their mutual interactions using a 3D Convolutional layer $\Psi_{3D}$:
\begin{equation}
\mathbf{F}_{geo} = \Psi_{3D}(\text{Concat}(\mathbf{F}_{map}, \mathbf{F}_{cam})).
\end{equation}

This fused control signal $\mathbf{F}_{geo}$ is then injected into the main DiT backbone via element-wise addition (ControlNet-like residuals) at the input level, enabling precise geometric guidance without disrupting the pre-trained generative priors.

\subsection{Curriculum Training with Progressive Masking} \label{sec:curriculum}
Given the inherent bias of standard video models towards text prompts, a naive training strategy often leads to the model neglecting geometric conditions (e.g., generating a generic "kitchen" while disregarding the specific navigation path). As recently emphasized in multimedia learning paradigms \cite{wang2024curriculum}, dynamically adjusting learning difficulty is crucial for complex multimodal tasks. To mitigate the modality bias, we introduce a curriculum training strategy that progressively controls the visibility of different input conditions.

Specifically, the training is divided into three progressive stages, namely warmup, annealing, and cooldown, with $S_{\text{warm}}$, $S_{\text{anneal}}$, and $S_{\text{cool}}$ steps allocated to each stage, respectively. And we define a normalized training progress variable $r(s) \in [0, 1]$ based on the training step index $s$. 
The progress function is formulated as:
\begin{equation}
r(s) =
\begin{cases}
0, & s < S_{\text{warm}}, \\[4pt]
\dfrac{s - S_{\text{warm}}}{S_{\text{anneal}}}, & S_{\text{warm}} \le s < S_{\text{warm}} + S_{\text{anneal}}, \\[6pt]
1, & s \ge S_{\text{total}} - S_{\text{cool}}.
\end{cases}
\end{equation}

To ensure a smooth transition, we apply a cosine kernel~\cite{nichol2021improveddenoisingdiffusionprobabilistic} to $r(s)$, formulated as $K_{cos}(r(s)) = \frac{1 - \cos(\pi \cdot r(s))}{2}$. With this schedule defined, the core principle of our strategy is to gradually reduce the model's reliance on auxiliary modalities, thereby compelling it to increasingly attend to the geometric conditions encoded in the BEV map.

\noindent
\textbf{Auxiliary Conditions Schedule.}
We apply step-dependent masking to auxiliary input conditions, including the camera trajectory $\boldsymbol{\tau}$, the reference image $\mathbf{x}_0$, and the instruction prompt $\mathbf{q}$. At each training step $s$, these conditions are independently masked with probability $P_c(s)$. For conditions $c \in \{\boldsymbol{\tau}, \mathbf{x}_0\}$, we adopt a standard cosine-scheduled masking probability:
\begin{equation}
    P_{c}(s) = P_{\text{init}}^{c} +
\big(P_{\text{end}}^{c} - P_{\text{init}}^{c}\big)\cdot K_{\text{cos}}(r(s)),
\end{equation}
which gradually increases the masking rate from an initial low probability $P_{\text{init}}^{c}$ to a final high probability $P_{\text{end}}^{c}$ as training progresses.

For the instruction condition $\mathbf{q}$, we empirically observe that models exhibit a strong dependency on it. To more strictly penalize this over-reliance, we apply a more aggressive masking strategy by warping the training progress:
\begin{equation}
    P_{q}(s) = P_{\text{init}}^{q} + (P_{\text{end}}^{q} - P_{\text{init}}^{q}) \cdot K_{\text{cos}}(r(s)^{0.5}).
\end{equation}

The term $r(s)^{0.5}$ ensures that the instruction masking rate increases significantly faster in the early stages compared to other conditions, effectively suppressing shortcut reliance on textual cues.

\noindent
\textbf{BEV map Schedule.}
The BEV map $\mathbf{m}$ is the core condition of our proposed task. Therefore, we maintain a constant, low dropout probability $P_{m}(s) = c$ throughout the training to maximize the model's exposure to this novel modality.

\noindent
\textbf{Inference.}
As detailed in Sec.~\ref{sec:task_setup}, during the inference stage, we evaluate our models under two distinct configurations: the full condition set and the BEV map-only mode.

\begin{table*}[t]
    \centering
    \renewcommand{\arraystretch}{1.3}
    \caption{
    \textbf{Quantitative comparison of our framework built upon different base models for the Map2Vid task.}
    The symbol $\circ$ denotes methods adapted by us.
    We report performance under two inference settings:
    ``All'' (utilizing the full condition set) and
    ``BEV'' (conditioned exclusively on the BEV map).
    The best and second-best results are highlighted
    with darker and lighter shades, respectively,
    using purple for visual quality and blue for
    geometric consistency.}
    \label{tab:comparison}
    \resizebox{0.95\linewidth}{!}{
        \setlength{\tabcolsep}{12pt}
        \begin{tabular}{cccccccc}
        \toprule
        \textbf{Method} &
        \textbf{Condition} &
        \textbf{FID} $\downarrow$ &
        \textbf{FVD} $\downarrow$ &
        \textbf{LPIPS} $\downarrow$ &
        \textbf{SFM-SR} $\uparrow$ &
        $\mathbf{APE}_{\text{trans}}$ $\downarrow$ &
        $\mathbf{APE}_{\text{rot}}$ $\downarrow$ \\
        \hline

        \multirow{2}{*}{Wan2.2-Camera}$\circ$
        & ALL
        & \cellcolor{blue!10} 47.64
        & \cellcolor{blue!20} 336.02
        & \cellcolor{blue!20} 0.538
        & \cellcolor{cyan!20} 58.7
        & \cellcolor{cyan!10} 1.93
        & 112.36 \\
        & BEV
        & 66.37
        & 457.32
        & 0.568
        & 51.3
        & 1.96
        & 116.29 \\
        \hline

        \multirow{2}{*}{HunyuanVideo V1.5}$\circ$
        & ALL
        & \cellcolor{blue!20} 42.78
        & 516.00
        & 0.589
        & \cellcolor{cyan!10} 56.6
        & \cellcolor{cyan!20} 1.73
        & \cellcolor{cyan!10} 108.79 \\
        & BEV
        & 196.27
        & 1255.75
        & 0.652
        & 23.7
        & 2.26
        & 122.32 \\
        \hline

        \multirow{2}{*}{Open-Sora V1.3}$\circ$
        & ALL
        & 57.08
        & 441.69
        & \cellcolor{blue!10} 0.539
        & 46.3
        & 2.11
        & 120.68 \\
        & BEV
        & 161.05
        & 1129.96
        & 0.612
        & 27.0
        & 2.46
        & 124.84 \\
        \hline

        \multirow{2}{*}{CogVideoX V1.5}$\circ$
        & ALL
        & 61.41
        & \cellcolor{blue!10} 411.82
        & 0.556
        & 55.3
        & 1.99
        & \cellcolor{cyan!20} 106.29 \\
        & BEV
        & 119.30
        & 1350.64
        & 0.586
        & 28.3
        & 2.07
        & 117.70 \\
        \hline
        \end{tabular}
    }
\end{table*}

\section{Experiments}
\label{sec:experiments}

\subsection{Experimental Setup}
We build our method on four DiT-based video generation models: \textit{Wan2.2-5B-Camera Control}~\cite{wan2025wan}, \textit{Hunyuan Video v1.5 (8.3B)}~\cite{wu2025hunyuanvideo}, \textit{CogVideoX v1.5 (5B)}~\cite{yang2024cogvideox}, and \textit{OpenSora v1.3 (1B)}~\cite{opensora}. Among them, Wan2.2-5B-Camera Control natively supports camera trajectory conditioning.
For curriculum training, we run the optimization for 2000 iterations, where the first  $10\%$ is used for warmup and the last $10\%$ for cooldown. The initial and final conditions masking probabilities are set to $P_{\text{init}}=0.1$ and $P_{\text{end}}=0.9$, respectively. All experiments are conducted on 8 NVIDIA H100 GPUs. For more details, please refer to the appendix.


\paragraph{Evaluation Metrics.}
We assess performance across two dimensions:
(1) \textbf{Visual Quality:} We report \textbf{FID}, \textbf{FVD}, and \textbf{LPIPS} to measure photorealism and temporal coherence.
(2) \textbf{Geometric Consistency:} To verify embodied navigation capabilities, we reconstruct 3D trajectories from generated videos using VGGSfM~\cite{wang2024vggsfm}. We measure the \textbf{SfM Success Rate (SfM-SR)}, \textbf{Absolute Pose Error (APE)}, and \textbf{Rotation Error (APE-Rot)} between the input and reconstructed trajectories.

\begin{table*}[t]
    \centering
    \renewcommand{\arraystretch}{1.3}
    \caption{
        \textbf{Ablation on inference conditions.}
        The best and second-best results are highlighted with
        darker and lighter shades, respectively, using blue
        for visual quality and cyan for geometric consistency.
    }
    \label{tab:ablation}
    \resizebox{0.95\linewidth}{!}{
        \setlength{\tabcolsep}{5pt}
        \begin{tabular}{cccccccccc}
        \hline
        \textbf{Instruction} &
        \textbf{First Frame} &
        \textbf{Trajectory} &
        \textbf{BEV Map} &
        \textbf{FID} $\downarrow$ &
        \textbf{FVD} $\downarrow$ &
        \textbf{LPIPS} $\downarrow$ &
        \textbf{SfM-SR} $\uparrow$ &
        $\mathbf{APE}_{\text{trans}}$ $\downarrow$ &
        $\mathbf{APE}_{\text{rot}}$ $\downarrow$ \\
        \hline

        - & - & \checkmark & - &
        86.38 & 856.80 & 0.568 & 33.3 & 2.10 & 120.28 \\

        - & - & - & \checkmark &
        66.37 &
        \cellcolor{blue!10}457.32 &
        0.568 &
        \cellcolor{cyan!10}51.3 &
        \cellcolor{cyan!10}1.96 &
        116.29 \\

        - & - & \checkmark & \checkmark &
        91.31 & 948.52 & 0.593 & 43.3 & 2.17 & 119.28 \\

        \checkmark & \checkmark & - & - &
        61.22 &
        673.08 &
        \cellcolor{blue!20}0.530 &
        35.0 & 2.37 & 123.80 \\

        \checkmark & \checkmark & \checkmark & - &
        62.92 &
        650.05 &
        \cellcolor{blue!10}0.532 &
        41.6 &
        2.25 &
        \cellcolor{cyan!10}112.70 \\

        \checkmark & \checkmark & - & \checkmark &
        \cellcolor{blue!10}60.41 &
        663.08 &
        \cellcolor{blue!20}0.530 &
        50.0 &
        2.37 &
        121.58 \\

        \checkmark & \checkmark & \checkmark & \checkmark &
        \cellcolor{blue!20}47.64 &
        \cellcolor{blue!20}336.02 &
        0.538 &
        \cellcolor{cyan!20}58.7 &
        \cellcolor{cyan!20}1.93 &
        \cellcolor{cyan!20}112.36 \\

        \hline
        \end{tabular}
    }
\end{table*}

\begin{figure*}[h!]
    \centering
    \includegraphics[width=\textwidth]{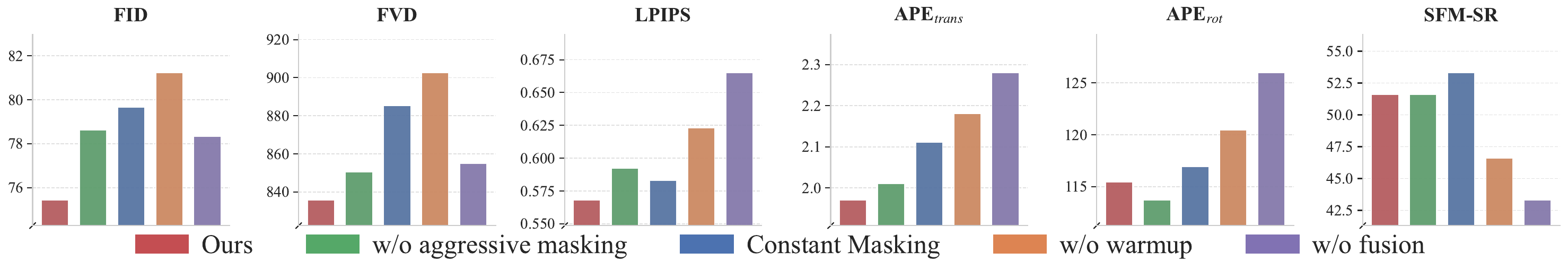}
    \caption{\textbf{Quantitative comparison of training strategies.} 
    We compare our method with four ablation variants: w/o Aggressive Masking, Constant Masking, w/o Warmup, and w/o Fusion. 
    The results demonstrate that our proposed curriculum strategy significantly improves geometric alignment without compromising visual quality compared to the baselines.}
    \label{fig:training_ablation}
\end{figure*}

\subsection{Main Results}
\label{sec:main_result}

We select 6 representative indoor scenes with substantial diversity including church, office, villa,
basement, hotel, and restaurant from the Map2Vid-100K dataset with a per-scene training protocol to enable controlled analysis, following common practice in simulation-based embodied settings\cite{adamkiewicz2022vision}. For evaluation, we generate 50 videos per scene per model under two inference configurations: full conditioning (All) and map-only conditioning (BEV). Tab.~\ref{tab:comparison} reports the quantitative results. We analyze these results from multiple perspectives below.

\noindent
\textbf{All vs. BEV Map-only Conditioning.} Inference with the full set of control signals consistently outperforms that relies solely on the BEV map across all backbones and metrics. This indicates that auxiliary conditions play a crucial role in stabilizing generation and improving both visual quality and temporal coherence. At the same time, it highlights the inherent difficulty of the Map2Vid task when rich semantic guidance and explicit low-level motion cues (camera trajectories) are absent. Notably, Wan exhibits the smallest performance drop among all backbones, with especially pronounced advantages on motion-related metrics (e.g., SfM-SR). A plausible explanation is Wan2.2-Camera's trajectory-based pretraining, which equips it with stronger motion and geometric priors and thus enables it to better leverage the control signals from the BEV map.

\noindent
\textbf{Model-wise Performance.} From a model-centric perspective, Wan and Hunyuan achieve the strongest overall performance across metrics. This result is consistent with their performance in general video generation, suggesting that the visual and temporal priors learned through large-scale generic video pretraining remain highly beneficial for embodied navigation video generation.

\subsection{Ablation Study}

\noindent
\textbf{Inference with Different Control Signals.} Following the same experimental setup as in Sec.~\ref{sec:main_result}, Tab.~\ref{tab:ablation} evaluates the effects of different combinations of control signals during the inference stage. Several observations emerge from the results: \ding{172} Full conditioning achieves the best performance on five out of six metrics, including both distributional visual metrics and all three geometric metrics, while yielding a slightly higher LPIPS than some reduced-condition configurations, indicating that semantic cues (instruction and first frame) and geometric cues (BEV map and trajectory) are heterogeneous geometric cues. \ding{173} Adding the BEV map and camera trajectory improves motion-related metrics. The BEV map is a stronger standalone control signal than the trajectory, suggesting that it provides richer and more stable global spatial constraints, whereas trajectory-only conditioning lacks sufficient context and is prone to geometric drift. \ding{174} Semantic cues are critical for perceptual quality. Removing the instruction and first frame leads to a pronounced degradation in perceptual metrics, highlighting their role as strong anchors for appearance and semantic consistency.

\noindent
\textbf{Model Architecture \& Training Strategy.} To isolate the effect of the training strategy, we conduct ablations on a single representative scene. For evaluation, 60 videos are generated for each setting. We consider the following variant training strategy: \textbf{(1) w/o Fusion}: BEV map and trajectory features are processed independently without 3D convolution; \textbf{(2) w/o Aggressive Masking}: text masking follows $r(s)$ instead of $r(s)^{0.5}$; \textbf{(3) Constant Masking}: masking probabilities are fixed ($0.9$ for auxiliary conditions and $0.1$ for the BEV map). \textbf{(4) w/o Warmup}: warmup and cooldown phases are removed. As shown in Fig.~\ref{fig:training_ablation}, we have several insights: \ding{172} Removing feature fusion (w/o Fusion) leads to a pronounced performance drop, particularly on motion metrics, indicating that jointly encoding the BEV map and camera trajectory is critical for capturing their strong geometric coupling. \ding{173} Variants with constant masking or without warmup consistently underperform, highlighting the importance of a smooth, progressive masking schedule for reducing reliance on auxiliary conditions. \ding{174} w/o Aggressive Masking variant confirms that aggressively suppressing textual cues helps prevent shortcut learning, this suggests that more principled training strategies could be a promising direction for future work.

\begin{table}[t]
    \centering
    \renewcommand{\arraystretch}{1.3}
    \caption{\textbf{Quantitative evaluation of zero-shot generalization capabilities for the Map2Vid task.} All evaluations are conducted using the fine-tuned Wan2.2-Camera model under the BEV-only conditioning setting.}
    \label{tab:zero_shot_generalization}
    \resizebox{\linewidth}{!}{
        \setlength{\tabcolsep}{12pt}
        \begin{tabular}{lccc}
        \toprule
        \textbf{Metric} & \textbf{Original} & \textbf{Cross-Scene} & \textbf{Sim2Real} \\ \midrule
        \textbf{FID} $\downarrow$ &  66.37 & 160.35 & 188.32 \\
        \textbf{FVD} $\downarrow$ &  457.32 & 2980.71 & 2319.96 \\
        \textbf{LPIPS} $\downarrow$ &  0.568 & 0.594     & 0.612 \\
        \textbf{SFM-SR} $\uparrow$ &  51.3 & 44.3 & 39.2 \\
        $\mathbf{APE}_{\text{trans}}$ $\downarrow$ &  1.96 & 4.36 & 5.25 \\
        $\mathbf{APE}_{\text{rot}}$ $\downarrow$ &  116.29 & 114.43 & 126.92 \\ \bottomrule
        \end{tabular}
    } 
\end{table}

\subsection{Zero-shot Generalization}
\label{sec:zero_shot_generalization}

To evaluate the model's generalization capabilities, we assess the fine-tuned Wan2.2-Camera model across three distinct settings: \textbf{Original} (trained and evaluated on the same simulated scene), \textbf{Cross-Scene} (trained on a specific scene and evaluated zero-shot on unseen simulated scenes), and \textbf{Sim2Real} (direct zero-shot transfer to real-world environments). All evaluations are conducted under the map-only (BEV-only) conditioning setting. Notably, for the Sim2Real setting, the input BEV maps are completely hand-drawn to verify the model's robustness against highly abstract and non-standard inputs.

\noindent
\textbf{Visual Quality Degradation.} As shown in Tab.~\ref{tab:zero_shot_generalization}, due to our scene-by-scene fine-tuning strategy, the visual quality metrics (FID, FVD, and LPIPS) suffer a severe degradation when the model is evaluated in zero-shot domains (Cross-Scene and Sim2Real) relying solely on the BEV map. Stripped of strong semantic anchors such as text instructions and reference frames, the model is forced into unconstrained texture hallucination in novel environments. This appearance ambiguity inevitably disrupts the spatiotemporal coherence of the generated videos.

\noindent
\textbf{Robust Geometric Control.} Despite the substantial decline in visual fidelity, the degradation in geometric metrics representing camera trajectory motion (SFM-SR, $\text{APE}_{\text{trans}}$, and $\text{APE}_{\text{rot}}$) is remarkably less pronounced. Even in the highly challenging Sim2Real setting---which combines completely hand-drawn BEV maps with a massive real-world domain gap---the model maintains considerable 3D structural integrity and trajectory alignment. This stark contrast between visual and geometric performance degradation strongly indicates that the model has not merely overfitted to the visual textures of the training scenes, but has genuinely learned the generalized ability to control long-horizon camera motion directly from the BEV map conditions.

\subsection{Human-Friendly Control Evaluation}

We conduct a user study involving
50 participants randomly recruited on campus to evaluate the human-friendliness of BEV maps as a motion
control representation. We randomly
sample five pairs of corresponding BEV maps and camera
trajectories from our dataset and present both representations
to the participants for comparison. Participants are asked
to indicate their preferred representation based on three
criteria: understandability, perceived editability, and
overall preference. As shown in Tab.~\ref{tab:user_study},
BEV maps receive 100\% preference across all three criteria.
These results demonstrate a clear preference for BEV maps
over raw camera trajectory parameters, supporting their
potential as an intuitive and human-friendly interface for
navigation motion control.

\begin{table}[t]
\caption{\textbf{User study on the human-friendliness of navigation
control representations.} Results report the percentage of
participants preferring each representation across three
usability criteria.}
\label{tab:user_study}
\centering

\renewcommand{\arraystretch}{1.3}

        \begin{tabular}{lcc}
        \toprule
        \textbf{Metric} &
        \textbf{Camera Trajectory} &
        \textbf{BEV Map} \\
        \midrule
        \textbf{Understandability} & 0\% & 100\% \\
        \textbf{Editability} & 0\% & 100\% \\
        \textbf{Overall Preference} & 0\% & 100\% \\
        \bottomrule
        \end{tabular}
    
\end{table}

\begin{table}[t]
    \caption{\textbf{Top-1 accuracy of VLM-based route verification
    using Map2Vid-generated videos.} Each instruction has three
    candidate BEV routes: one ground-truth route and two
    perturbations. Gain is measured relative to random
    selection (33.3\%).}
    \label{tab:route_ranking}
    \centering
    \renewcommand{\arraystretch}{1.3}
    \begin{tabular}{lcc}
        \toprule
        \textbf{VLM} &
        \textbf{Top-1 Acc. (\%) $\uparrow$} &
        \textbf{Gain (pp)} \\
        \midrule
        \textbf{Qwen3} & \textbf{50.0} & \textbf{+16.7} \\
        \textbf{ChatGPT-5.5} & \textbf{70.0} & \textbf{+36.7} \\
        \textbf{Gemini-3.1 Pro} & \textbf{60.0} & \textbf{+26.7} \\
        \textbf{Random Guess} & \textbf{33.3} & \textbf{--} \\
        \bottomrule
    \end{tabular}
\end{table}

\begin{figure}[t]
    \centering
    \includegraphics[width=\linewidth]{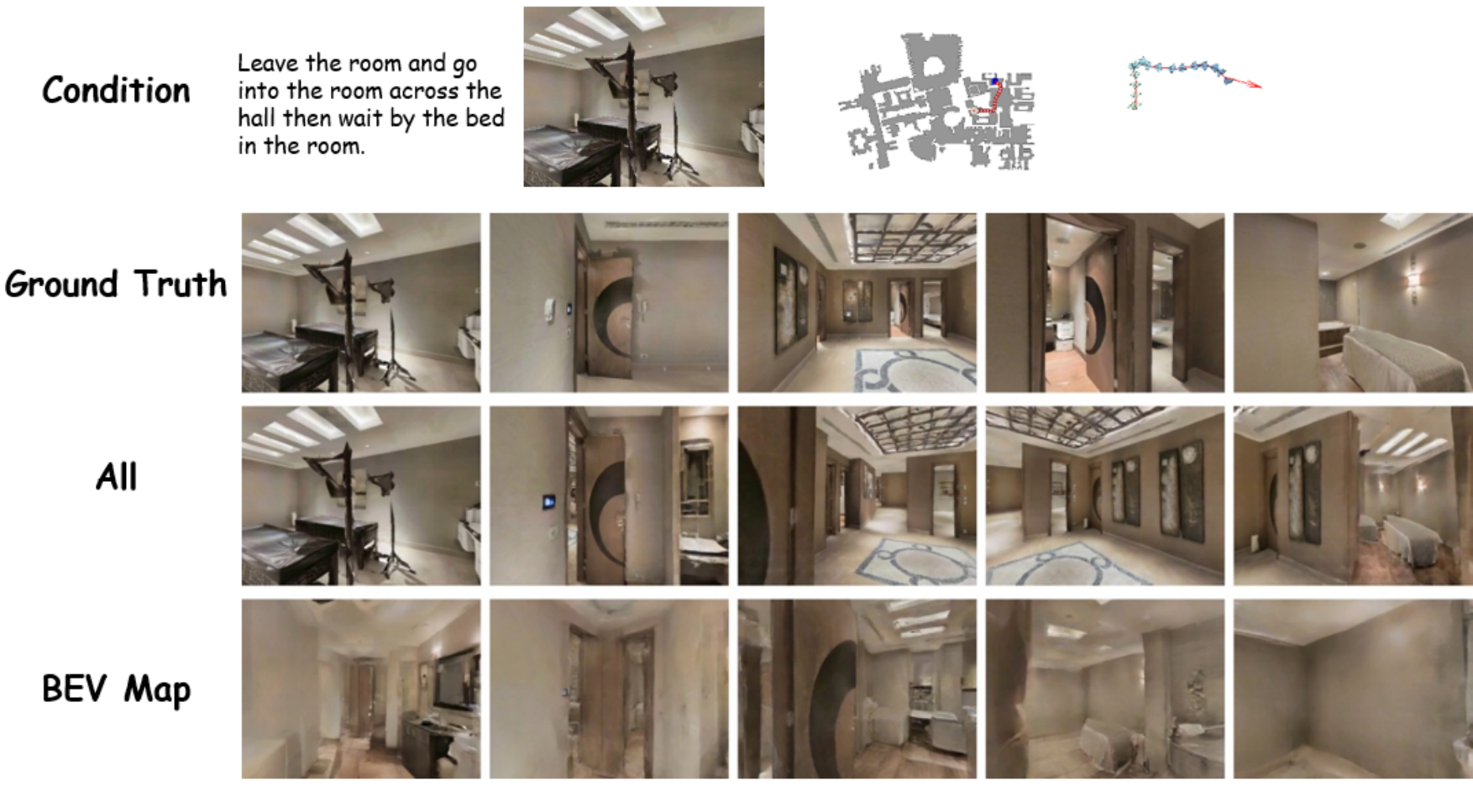}
    \caption{\textbf{Qualitative results.} From top to bottom: ground-truth video, and generated results via full conditions and BEV map, respectively.
    }
    \vspace{-0.15in}
    \label{fig:viz_result}
\end{figure}

\begin{figure}[t]
    \centering
    \includegraphics[width=0.45\linewidth]{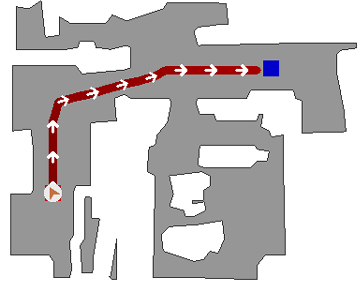}
    \includegraphics[width=0.45\linewidth]{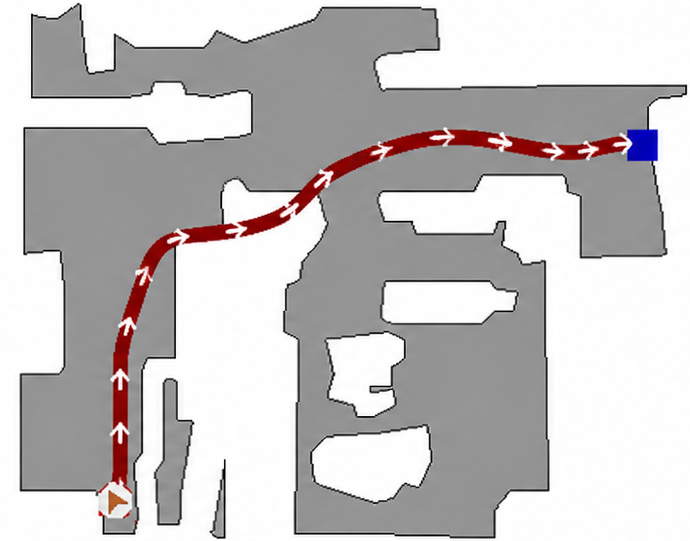}
    \caption{\textbf{Qualitative comparison between the ground-truth
    BEV route(left) and the trajectory reconstructed from a
    Map2Vid-generated video using VGGSfM(right). }The recovered route
    follows the overall navigation structure with local deviations.}
\label{fig:route_alignment}
\end{figure}

\subsection{Potential Application to Vision-and-Language Navigation}

Beyond controllable video generation, Map2Vid has the
potential to support vision-and-language navigation (VLN)
by providing visual imagination of candidate navigation
routes. Specifically, the generated egocentric videos
allow navigation agents to anticipate visual observations
along different routes before execution, potentially
facilitating more informed route selection. To explore
this potential utility, we conduct a proof-of-concept
experiment in which a VLM selects the most
instruction-consistent route based on generated videos.

For each navigation instruction, we construct three
candidate BEV routes, including the ground-truth route
and two perturbed routes with incorrect turns or shifted
endpoints. Map2Vid generates an egocentric video for each
candidate, and the VLM selects the video that best matches
the instruction. We evaluate 10 such instances, each
containing three candidate videos. As shown in
Tab.~\ref{tab:route_ranking}, the evaluated VLMs achieve
top-1 accuracies of 50.0\%, 70.0\%, and 60.0\%,
respectively, exceeding the random baseline of 33.3\%.
Although preliminary, these results demonstrate the
feasibility of leveraging Map2Vid-generated videos as
visual evidence for candidate-route selection, highlighting
their potential to assist downstream VLN decision-making.
Beyond route selection, the generated videos could also
support future observation prediction, look-ahead
navigation planning, and synthetic data augmentation
for navigation policy learning. We leave the systematic
integration of Map2Vid into VLN agents and the evaluation
of its impact on downstream navigation performance
to future work.

\subsection{Qualitative Results}

\noindent\textbf{Qualitative results.} Fig.~\ref{fig:viz_result} shows qualitative results generated by Wan2.2-Camera. Overall, both \emph{full-conditioning} and \emph{map-only} inference produce long-horizon navigation videos with substantial egocentric motion while maintaining reasonable per-frame visual quality, without obvious artifacts such as excessive blur, temporal flickering, or physically implausible distortions. Without semantic guidance from the reference frame and language instruction, map-only generations often start from an appearance that deviates from the ground-truth, but this discrepancy tends to diminish as the video progresses. In terms of path following, both settings recover the overall trend of camera motion, though fine-grained motion remains challenging, with errors such as under- or over-rotation at turning points and temporal drift in translation.
More results can be found in the Appendix.

\noindent\textbf{Trajectory Following Visualization.}
To further assess the geometric controllability of Map2Vid, we visualize the camera trajectories reconstructed from generated videos and compare them with the corresponding ground-truth navigation routes on BEV maps. Specifically, we employ VGGSfM to recover the camera motion from generated egocentric videos and align the reconstructed trajectories with the reference coordinate system. As illustrated in Fig.~\ref{fig:route_alignment}, the recovered trajectory in this example closely follows the specified navigation route, preserving its overall motion pattern and turning structure. This qualitative observation complements the trajectory accuracy metrics and provides additional evidence that BEV maps can guide egocentric video generation along designated navigation routes.

\section{Limitations and Future Work}
\label{sec:limitations}

While Map2Vid establishes a robust baseline for map-conditioned navigation video generation, several limitations warrant further investigation. 
Our study is conducted in a controlled simulated setting. The current benchmark is built by replaying R2R trajectories in Habitat over Matterport3D scenes, which enables precise synchronization between routed BEV maps, egocentric videos, language instructions, and camera trajectories. This design is important for task formulation and reproducible evaluation, but it also means that our conclusions are limited to a simulation-based benchmark rather than real-world deployment. In particular, appearance statistics, rendering artifacts, and motion dynamics in simulation may differ from those encountered by physical embodied agents. A natural next step is therefore to extend Map2Vid toward real-world or mixed-reality data sources, and to study whether BEV-based control remains effective under stronger domain variability.

Furthermore, while the framework effectively captures global navigation intent, fine-grained motion execution, such as precise angular orientation at sharp turns or subtle translational stability, remains a challenge, as evidenced by occasional errors in trajectory adherence observed in our qualitative analysis. Another constraint lies in the static nature of the current dataset: our formulation primarily focuses on fixed indoor layouts and does not yet account for dynamic obstacles or human-agent interactions, which are critical for comprehensive embodied intelligence. Finally, the reliance on large-scale Diffusion Transformers (DiTs) such as Wan2.2 and HunyuanVideo involves substantial computational overhead and inference latency, suggesting that future research into model distillation or efficient temporal attention mechanisms is necessary.~\cite{Mao2026LDT}

\section{Conclusion}

In this work, we present \textit{Map2Vid}, a novel framework that addresses the limitations of existing video generation models in providing intuitive and spatially grounded control for embodied navigation. By introducing Bird's-Eye-View (BEV) maps as a human-friendly geometric condition, Map2Vid enables the synthesis of coherent egocentric videos that respect global scene layouts. A curriculum training strategy with progressive masking is adopted to regulate condition visibility, preventing shortcut learning and ensuring the model effectively internalizes the spatial dynamics encoded in the BEV maps. Through the construction of the large-scale Map2Vid-100K benchmark, it is found that our method supports flexible inference, achieving robust performance even when conditioned solely on BEV maps without explicit camera trajectories. Experimental results across multiple foundation models demonstrate the effectiveness and promise of the Map2Vid direction.

\bibliographystyle{IEEEtran}
\bibliography{acmart}

\appendix

In this appendix, we provide supplementary materials to complement the main text. Specifically, we first detail the implementation settings, encompassing the two-stage dataset construction process—which includes floor segmentation and trajectory verification—as well as comprehensive model and training configurations across various architectures. We then present further ablation studies analyzing the underlying mechanisms of our framework, particularly focusing on how the synergy between semantic and geometric streams shapes the generation process under the Full Control setting. Finally, we provide additional qualitative results that compare different conditioning protocols to empirically validate the efficacy of Bird's-Eye-View (BEV) maps as a primary control signal for controllable navigation video generation. 

\section{Implementation Details}
\label{sec:implementation_details}

\subsection{Dataset Construction}

Modeling navigation across different floors introduces complex vertical transitions and spatial discontinuities, which are notoriously difficult for generative models to synthesize accurately. To ensure data consistency and focus on high-fidelity planar navigation, we strictly exclude trajectories involving cross-floor actions. Specifically, we employ a two-stage filtering mechanism consisting of floor segmentation and trajectory verification.
\begin{itemize}
    \item \textbf{Floor Segmentation.} We first initialize a floor detector that performs a bottom-up vertical scanning of the entire 3D scene. By analyzing the projected top-down views of the navigation mesh at varying altitudes, the detector computes pixel similarity scores between adjacent vertical slices. Significant deviations in similarity allow us to identify floor boundaries, thereby defining precise height intervals (coordinates along the $Y$-axis) for each distinct floor.
    \item \textbf{Trajectory Verification.} For each candidate agent movement task, the system pre-loads the coordinates of all trajectory points. We utilize the pre-computed height interval database to assign a floor index to each point. The system iterates through the trajectory; if any two consecutive valid path points are detected to belong to different floor indices, the task is classified as a multi-floor action. Consequently, such tasks are marked as invalid and excluded from the final video generation pipeline.
\end{itemize}

\begin{table*}[t!]
    \centering
    \renewcommand{\arraystretch}{1.3} 
    \caption{\textbf{Quantitative evaluation of ablation studies on model architecture and training strategies.} Complementing the qualitative comparison in Fig. 4 of the main text, this table details the performance of different variants under both full control (ALL) and map-only (BEV) settings. The results demonstrate that our proposed geometric fusion and curriculum learning strategy achieve superior geometric consistency (SFM-SR, APE) while maintaining high visual quality, validating the effectiveness of these components.}
    \label{tab:training_ablation}
    \resizebox{0.95\linewidth}{!}{
        \setlength{\tabcolsep}{8pt}

        \begin{tabular}{cccccccc}
        \hline

        \textbf{Method} & \textbf{Condition} & \textbf{FID} $\downarrow$ & \textbf{FVD} $\downarrow$ & \textbf{LPIPS} $\downarrow$ & \textbf{SFM-SR} $\uparrow$ & $\mathbf{APE}_{\text{trans}}$ $\downarrow$ & $\mathbf{APE}_{\text{rot}}$ $\downarrow$ \\ \hline
        
        \multirow{2}{*}{Constant Masking}  & ALL     & 69.38  & 737.83  & 0.566  & 61.6  & 2.32 & 121.48 \\
                                           & BEV & 79.65  & 885.15  & 0.583  & 53.3  & 2.11 & 116.92 \\ \hline
        
        \multirow{2}{*}{w/o Aggressive Masking}      & ALL     & 63.04  & 660.11  & 0.559  & 58.7   & 2.16   & 119.34   \\
                                           & BEV & 78.62 & 850.32 & 0.592  & 51.3   & 2.01   & 113.69   \\ \hline
        
        \multirow{2}{*}{w/o warmup}        & ALL     & 68.32  & 701.85  & 0.578  & 56.6   & 2.11  & 117.42   \\
                                           & BEV & 81.22 & 902.52 & 0.623  & 46.6   & 2.18   & 120.46   \\ \hline

        \multirow{2}{*}{w/o fusion}        & ALL     & 71.32  & 742.89  & 0.593  & 48.3   & 2.21   & 122.63   \\
                                           & BEV & 78.33 & 854.92 & 0.665  & 43.3   & 2.28   & 125.97   \\ \hline
        
        \multirow{2}{*}{\textbf{Ours}}     & ALL     & 62.97  & 657.73  & 0.532  & 58.3   & 1.93 & 112.36  \\
                                           & BEV & \cellcolor{blue!1} 75.43 & 835.65 & 0.568  & 51.6   & 1.96 &  115.43   \\ \hline
        \end{tabular}
    }

\end{table*}

\noindent
\textbf{Visual Representation.} In the resulting dataset, navigation episodes are visualized using Bird's-Eye-View (BEV) maps. The agent's initial state is denoted by a white circle, with an inscribed arrow indicating its starting orientation. The traversal path is rendered as a red line, terminating at a destination marker represented by a blue square.

\subsection{Model \& Training Details}
\label{subsec:implementation}

To demonstrate the versatility of our proposed framework, we integrate the conditioning module into four representative Diffusion Transformer (DiT) architectures: Wan2.2-5B, HunyuanVideo v1.5, CogVideoX v1.5, and OpenSora v1.3.

\begin{itemize}
    \item \textbf{Data Preprocessing.} We standardize all video frames and their corresponding Plücker coordinates to a unified resolution of $384 \times 672$ via bilinear interpolation and center cropping. Pixel values are normalized to $[-1, 1]$ (mean and standard deviation of 0.5), whereas the Plücker coordinates remain unnormalized to preserve geometric fidelity.
    
    \item \textbf{Optimization.} We utilize the Adam optimizer across all experiments. For Wan2.2, HunyuanVideo, and CogVideoX, the learning rate is initialized at $2 \times 10^{-5}$ and follows a cosine decay schedule. In contrast, OpenSora is fine-tuned with a constant learning rate of $1 \times 10^{-4}$.
    
    \item \textbf{Hardware \& Training.} All models are trained on a cluster of 8 NVIDIA H100 (80GB) GPUs. We employ a global batch size ranging from 8 to 16, subject to memory constraints. The training process typically converges within 2,000 iterations.
\end{itemize}

\section{More Ablation Study On Model Architecture \& Training Strategy}
\label{sec:appendix_ablation}

In this section, we provide a qualitative analysis of the underlying mechanisms governing the Map2Vid framework. We specifically focus on the \textbf{Full Control} setting (incorporating BEV Map, Camera Trajectory, Text Instruction, and First Frame), analyzing how the synergy between semantic and geometric streams shapes the generation process compared to single-modality baselines.

\subsection{ Holistic Alignment via Full Control}

The quantitative superiority of the Full Control setting, particularly in perceptual metrics (FID, FVD), can be attributed to its ability to simultaneously constrain two manifolds in the latent space:

\noindent
\textbf{Appearance Anchoring.}
In the map-only setting, the model is tasked with hallucinating 3D textures solely from a schematic layout. While the BEV map effectively dictates the structural topology (e.g., wall placements, room connections), it remains texturally ambiguous. This ambiguity forces the generative backbone to sample from a high-entropy distribution of possible appearances, often leading to temporal flickering and inconsistent texturing as the video progresses.
The Full Control setting mitigates this by utilizing the \textit{First Frame} and \textit{Text Instruction} as strong semantic anchors. These inputs lock the generation into a specific subspace of the appearance manifold, resolving textural ambiguity. Consequently, the model can dedicate its geometric capacity—derived from the BEV map—to motion synthesis rather than texture hallucination.

\noindent
\textbf{Geometric Grounding.}
Conversely, relying exclusively on semantic cues (common in standard T2V models) often results in ``plausible but incorrect'' navigation, where agents may traverse obstacles or violate spatial continuity. By integrating the BEV map and camera trajectory within the Full Control stack, we impose strict boundary conditions on the agent's action space. Thus, Full Control does not merely provide \textit{more} information; it creates a complementary constraint system where semantic priors ensure visual fidelity while geometric priors enforce physical plausibility.


\section{More Qualitative Results}
\label{app:viz_results}

In this section, we provide more qualitative results to empirically validate the core premise of our proposed \textbf{Map2Vid} task: that BEV maps can serve as an effective, human-centric interface for controllable navigation video generation.

\begin{figure*}[!p]
    \centering
    \includegraphics[width=0.7\linewidth]{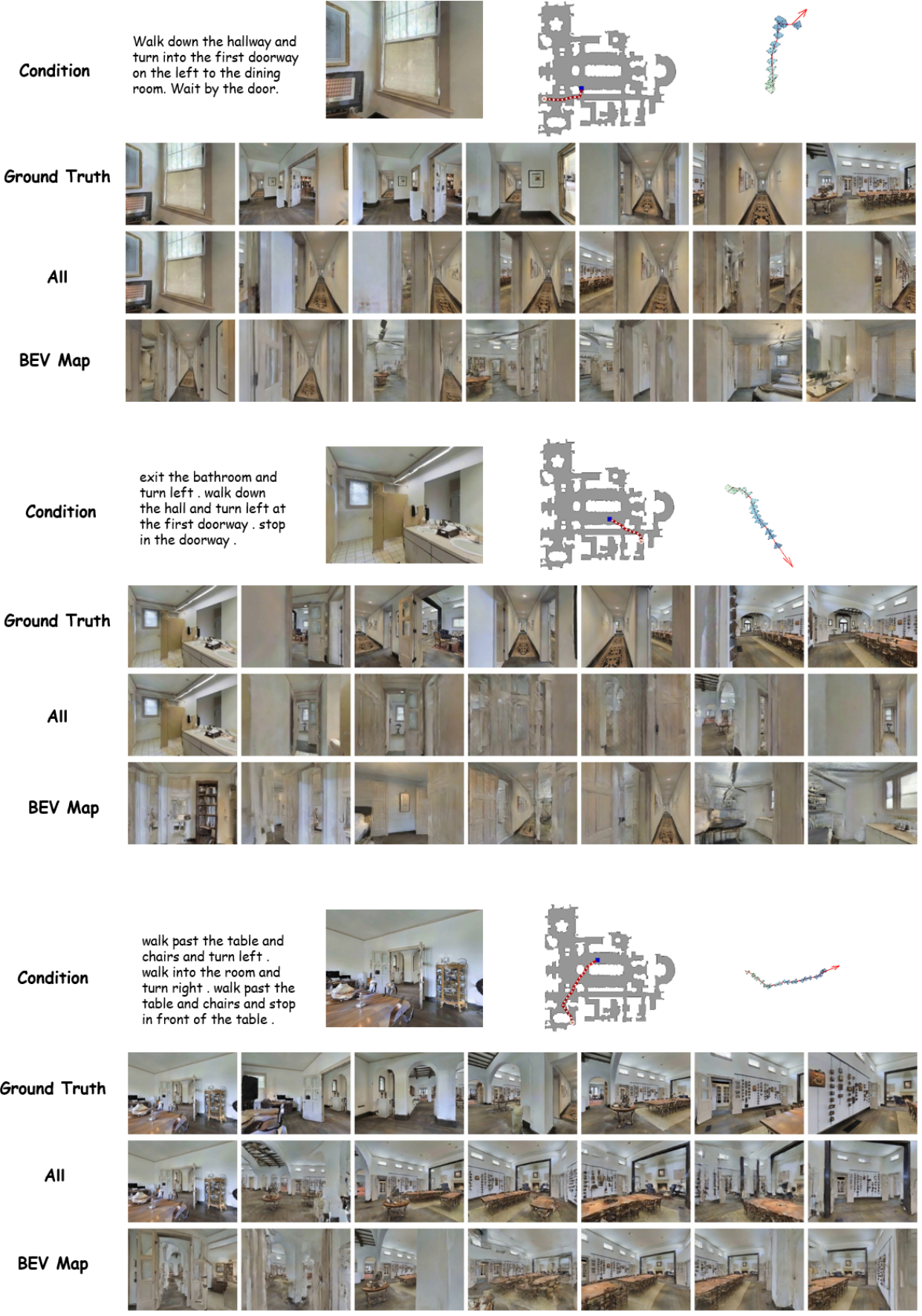}
    \caption{\textbf{Qualitative comparison under different control conditions.} We visualize results for three different scenes. In each group, the first column displays the input \textbf{Conditions}, and the second column shows the \textbf{Ground Truth} frames. The third column (\textbf{All}) presents the generation results conditioned on both the BEV Map and camera trajectory, while the fourth column (\textbf{BEV Map}) shows the results conditioned only on the BEV Map.}
    \label{fig:appendix_2n8kARJN3HM}
\end{figure*}

\begin{figure*}[ht!p]
    \centering
    \includegraphics[width=0.7\linewidth]{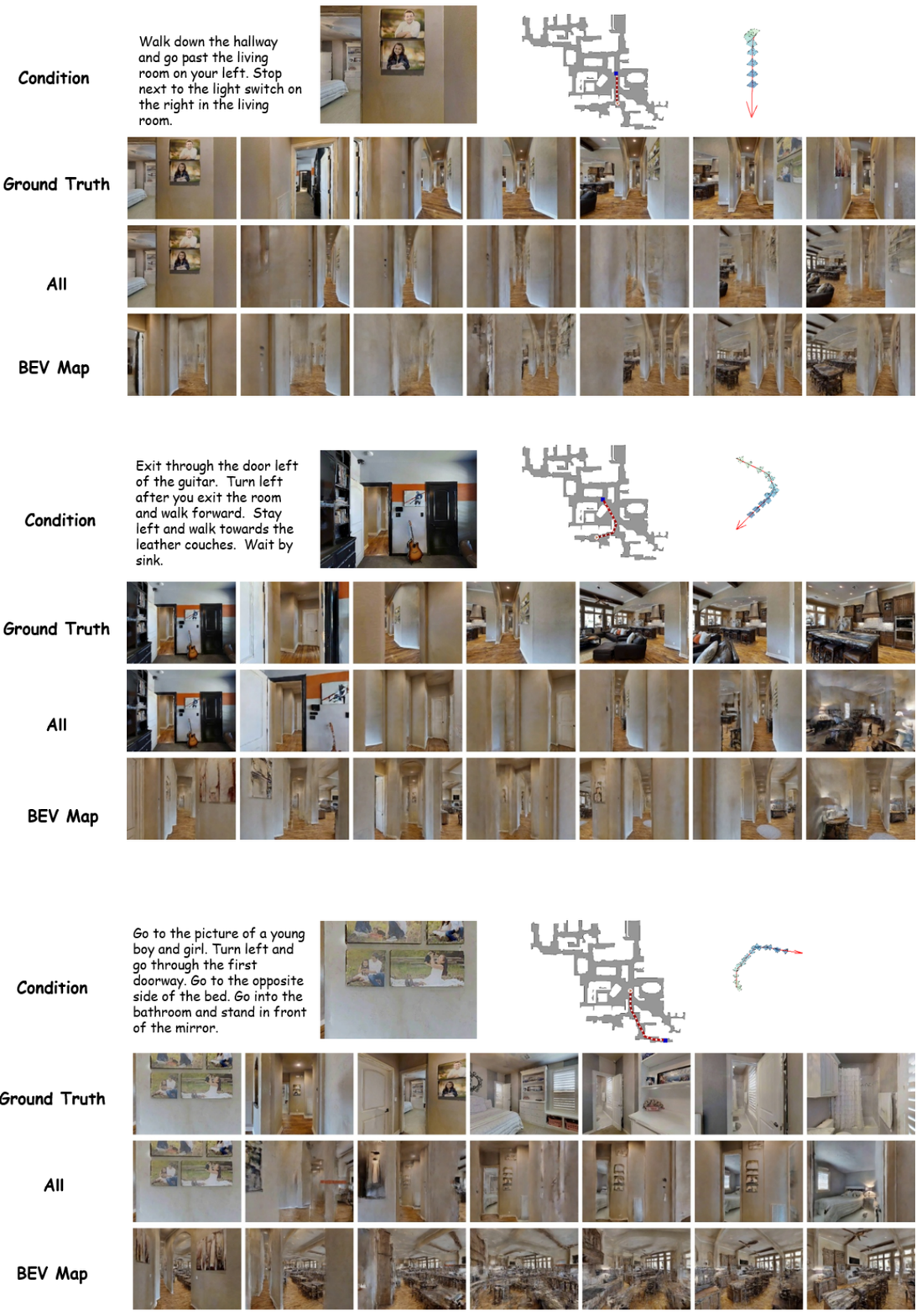}
    \caption{\textbf{Qualitative comparison under different control conditions.} }
    \label{fig:appendix_8WUmhLawc2A}
\end{figure*}

Fig~\ref{fig:appendix_2n8kARJN3HM}~\ref{fig:appendix_8WUmhLawc2A} visualizes generated navigation sequences across three diverse indoor environments. The corresponding video files for these figures are provided in the \texttt{examples} directory of the supplementary materials.Please note that some videos have been compressed to meet the file size constraints of the supplementary materials. To assess the efficacy of the BEV map as a primary control signal, we compare the Ground Truth videos against samples generated under two distinct conditioning protocols:

\begin{itemize}
    \item \textbf{Full Control (Reference):} The model operates with access to all auxiliary signals, including the \textit{First Frame}, \textit{Textual Instruction}, \textit{Camera Trajectory}, and the \textit{BEV Map}. This setting serves as an upper bound, verifying the model's capability to fuse multimodal inputs for precise reconstruction.
    \item \textbf{Map-only (Map2Vid Setting):} The model generates videos conditioned \textit{solely} on the \textit{BEV Map}, masking out low-level camera trajectories and initial visual cues. This setting directly evaluates the feasibility of our proposed task, testing whether the model can hallucinate 3D egocentric visual dynamics purely from 2D schematic layouts.
\end{itemize}

As illustrated in Fig.~\ref{fig:appendix_2n8kARJN3HM}~\ref{fig:appendix_8WUmhLawc2A}, the results under the \textbf{Map-only} setting are particularly noteworthy. Despite lacking explicit camera pose matrices or starting visual context, the model successfully synthesizes navigation videos that are semantically coherent. While the \textbf{Full Control} outputs (middle rows) naturally achieve higher fidelity in texture details due to the sexampltrong guidance from the first frame and trajectory, the \textbf{Map-only} outputs (bottom rows) correctly reflect the scene geometry, room transitions, and navigational intent. This qualitative evidence strongly supports the viability of Map2Vid, confirming that BEV maps provide adequate spatial grounding to guide the generation of long-horizon embodied navigation videos without the need for unintuitive low-level trajectory specifications.

\subsection{The Optimization Paradox in Full Control}

Although Full Control provides richer conditioning signals,
our ablation study reveals a potential optimization challenge:
\textbf{Modality Competency Collapse}.

DiT backbones pre-trained on large-scale video data tend to
favor familiar text and image conditions. Under naive
training (e.g., Constant Masking), the model may over-rely
on these semantic cues while underutilizing the novel BEV
signal, producing visually plausible videos with inaccurate
navigation paths.

Our curriculum masking strategy mitigates this issue by
temporarily suppressing dominant semantic conditions,
encouraging the model to learn BEV-based geometric
constraints before integrating all modalities.
This facilitates more effective multimodal conditioning
and improves navigation consistency.

\end{document}